# Credit Default Mining Using Combined Machine Learning and Heuristic Approach

Sheikh Rabiul Islam
*Computer Science*
Tennessee Technological University
Cookeville, U.S.A
sislam42@students.tntech.edu

William Eberle
*Computer Science*
Tennessee Technological University
Cookeville, U.S.A
weberle@tntech.edu

Sheikh Khaled Ghafoor
*Computer Science*
Tennessee Technological University
Cookeville, U.S.A
sghafoor@tntech.edu

*Abstract*—Predicting potential credit default accounts in advance is challenging. Traditional statistical techniques typically cannot handle large amounts of data and the dynamic nature of fraud and humans. To tackle this problem, recent research has focused on artificial and computational intelligence based approaches. In this work, we present and validate a heuristic approach to mine potential default accounts in advance where a risk probability is precomputed from all previous data and the risk probability for recent transactions are computed as soon they happen. Beside our heuristic approach, we also apply a recently proposed machine learning approach that has not been applied previously on our targeted dataset [15]. As a result, we find that these applied approaches outperform existing state-of-the-art approaches.

## I. INTRODUCTION

In general, we can refer to a customer's inability to pay, or their default on a payment, or personal bankruptcy, all as potential issues of non-payment. However, each of these scenarios is a result of different circumstances. Sometimes it is due to a sudden change in a person's income source due to job loss, health issues, or an inability to work. Sometimes it is a deliberate, for instance, when the customer knows that he/she is not solvent enough to use a credit card anymore, but still uses it until the card is stopped by the bank. In the latter case, it is a type of fraud, which is very difficult to predict, and a big issue to creditors.

To address this issue, credit card companies try to predict potential default, or assess the risk probability, on a payment in advance. From the creditor's side, the earlier the potential default accounts are detected the lower the losses [5]. For this reason, an effective approach for predicting a potential default account in advance is crucial for the creditors if they want to take preventive actions. In addition, they could also investigate and help the customer by providing necessary suggestions to avoid bankruptcy and minimize the loss.

Analyzing millions of transactions and making a prediction based on that is time consuming, resource intensive, and some time error prone due to the dynamic variables (e.g., balance limit, income, credit score, economic conditions, etc.). Thus, there is a need for optimal approaches that can deal with the above constraints. In our previous work [3], we proposed an approach that precomputes all previous data (offline data) and calculates a score. Subsequently, it waits for a new transaction (online data) to occur and calculate another score as soon as the transaction occurs. Finally, all scores are combined to make a decision. We used the term OLAP data for *offline* data and OLTP data for *online* data in our previous work [3].The main limitations of the previous work was the use of a synthetic dataset and a lack of validation of the proposed model using a publicly available, real-world dataset. Online Analytical Processing (OLAP) systems typically use archived historical data over several years from a data warehouse to gather business intelligence for decision-making and forecasting. On the other hand, Online Transaction Processing (OLTP) systems, only analyze records within a short window of recent activities - enough to successfully meet the requirement of current transactions within a reasonable response time [19][3].

Currently, a variety of Machine Learning approaches are used to detect fraud and predict payment defaults. Some of the more common techniques include K Nearest Neighbor, Support Vector Machine, Random Forest, Artificial Immune System, Meta-Learning, Ontology Graph, Genetic Algorithms, and Ensemble approaches. However, a potential approach that has not been used frequently in this area is *Extremely Random Trees,* or *Extremely Randomized Trees (ET)* [20]. This approach came about in 2006, and is a tree-based ensemble method for supervised classification and regression problems. In *Extremely Random Trees* (ET) randomness goes further than the randomness in Random Forest. In Random Forest, the splitting attribute is determined by some criteria where the attribute is the best to split on that level, whereas in ET the splitting attribute is also chosen in an extremely random manner in terms of both variable index and splitting value. In the extreme case, this algorithm randomly picks a single attribute and cut-point at each node, which leads to a totally randomized trees whose structures are independent of the target variables values in the learning sample [20]. Moreover, in ET, the whole training set is applied to train the tree instead of using bagging to produce the training set as in Random Forest. As a result, ET gives a better result than Random Forest for a particular set of problems. Besides accuracy, the main strength of the ET algorithm is its computational efficiency and robustness [20]. While ET does reduce the variance at the expense of an increase in bias, we will use this algorithm as the foundation for our proposed approach.

The following sections discuss related research, followed by our proposed approach. We then present the data that will be used, and our experimental results. We then conclude with some observations and future work.

## II. LITERATURE REVIEW

The research work of [1][2][3][4][5] are all about personal bankruptcy or credit card default on payment prediction and detection. In the work of [4], the authors worked on finding financial distress from four different summarized credit datasets. Bankruptcy prediction and credit scoring were the primary indicators of financial distress prediction. According to the authors, a single classifier is not good enough for a classification problem of this type. So they present an ensemble approach where multiple classifiers are used on the same problem and then the result from all classifiers are combined to get the final result, and reduce Type I/II errors – crucial in the financial sector. For their classification ensemble approach, they use four approaches: a) majority voting b) bagging c) boosting and 3) stacking. The also introduced a new approach called Unanimous Voting (UV) where if any of the classifiers says "yes" then it is assumed as "yes" whereas in Majority Voting (MV) at least $(n+1)/2$ classifiers need to say "yes" to make the final prediction yes. In the end, they are able to reduce the Type II error but decrease the overall accuracy.

In the work of [5], the authors present a system to predict personal bankruptcy by mining credit card data. In their application, each original attribute is transformed either as: i) a binary [good behavior and bad behavior] categorical attribute, or ii) a multivalued ordinal [good behavior and graded bad behavior] attribute. Consequently, they obtain two types of sequences, i.e., binary sequences and ordinal sequences. Later they use a clustering technique for discovering useful patterns that can help them to identify bad accounts from good accounts. Their system performs well, however, they only use a single source of data, whereas the bankruptcy prediction systems of credit bureaus use multiple data sources related to creditworthiness.

In the work of [1], they compared the accuracy of different data mining techniques for predicting the credit card defaulters. The dataset used in this research is from the UCI machine learning repository which is based on *Taiwan's* credit card clients default cases [15]. This dataset has 30,000 instances, and 6626 (22.1%) of these records are default cases. There are 23 features in this dataset. Some of the features include credit limit, gender, marital status, last 6 months bills, last 6 months payments, etc. These are labeled data and labeled with 0 (refer to non-default) or 1 (refers to default). From the experiment, based on the *area ratio in the lift chart* on the validation data, they ranked the algorithms as follows: artificial neural network, classification trees, naïve Bayesian classifiers, K-nearest neighbor classifiers, logistic regression, and discriminant analysis. In terms of accuracy, K-nearest neighbor demonstrated the best performance with an accuracy of 82% on the training data and 84% on the validation or test data. To get an actual probability of "default" (rather than just a discrete binary result) they proposed a novel approach called Sorting Smoothing Method (SSM).

In the work of [2], the authors use the same *Taiwan* dataset [15] as of [1]. However, they applied a different set of algorithms and approaches. In this research, they proposed an application of online learning for a credit card default detection system that achieves real-time model tuning with minimal efforts for computations. They mentioned that most of the available techniques in this area are based on offline machine learning techniques. Their work is the first work in this area that is capable of updating a model based on new data in real time. On the other hand, traditional algorithms require retraining the model even if there is some new data, and the size of the data affects the computation time, storage and processing. For the purpose of real-time model updating, they use Online Sequential Extreme Learning Machine (OS-ELM) and Online Adaptive Boosting (Online AdaBoost) methods in their experiment. They compared the results from above mentioned two algorithms with basic ELM and AdaBoost in terms of training efficiency and testing accuracy. In online AdaBoost, the weight for each weak leaner and the weight for the new data is updated based on the error rate found in each of the iterations. The OS-ELM is based on basic ELM which is formed from a single layer feedforward network. Along with these algorithms, they also applied some other classic algorithms such as KNN, SVM, RF, and NB. Although KNN, SVM, and RF have shown higher accuracy, the training time was more than 100 times compared to other algorithms. They found RF exhibits great performance in terms of efficiency and accuracy (81.96%). In the end, both online ELM and AdaBoost maintain the accuracy level of other offline algorithms, while significantly reducing the training time with an improvement of 99% percent. They conclude that the online AdaBoost has the best computational efficiency, and the offline or classic RF has best predictive accuracy. In other words, Online AdaBoost balances relatively better than offline or classic RF between computational accuracy and computational speed. They mentioned two future directions of this research as follows: a) incorporating concept drift to deal with the change of new data distributions over time, which may affect the effectiveness of the online learning model, and b) sustaining the robustness of online learning for a dataset with missing records or noise. They also mention that some other online learning techniques like Adaptive Bagging could be applied and compared in terms of speed, accuracy, stability, and robustness.

Besides credit card default prediction and detection, there are lots of work on different types of credit card fraud detection. Some of those are [6], [7], [8], [9], [10], [11], [12], [13], and [14], where credit card transaction fraud detection are emphasized and surveyed. Most of the transaction fraud is the direct result of stolen credit card information. Some of the techniques they used for credit card transaction fraud detection are as follows: Artificial Immune System, Meta-Learning, Ontology Graph, Genetic Algorithms, etc.

So, despite the plethora of research being done in the area of credit default/fraud detection, little has been reported that resolves the issue of detecting default/fraud *early* in the process. In this work, we will focus on detecting default accounts in the very early stage of credit analysis towards the discovery of a potential default on payment or even bankruptcy.

## III. METHODOLOGY

We applied two different approaches to the dataset: one for comparing results with previous standard machine learning approaches, and the other for validating our proposed approach. The first approach is the application of different machine learning algorithms on the dataset. We call this standard approach the *Machine Learning Approach*. The second approach is based on our previous work [3], where two tests are performed, what we call a *standard test* and a *customer specific test*, to mine potential defaulting accounts. We will call this second approach our *Heuristic Approach*.

The *Heuristic Approach* predicts the credit default risk in two steps. In the first step, we compute a credit default probability score from archived transaction history using appropriate machine learning algorithms. This score is stored in the database and continuously updated as new transactions occur. In the second step, as real-time transactions occur, we apply a heuristic (applying a standard test and a customer specific test, explained in detail in section V) to compute a risk score. This score is combined with the archived score using the equations (1) and (2) to compute overall risk probability.

For the *Machine Learning Approach*, we experimented with various supervised machine learning algorithms to determine the best algorithm. Then, for the *Heuristic* approach we will take the best algorithm found from the *Machine Learning Approach* and apply it only to the *offline* data to calculate the offline risk probability $R_{offline}$. And whenever a new transaction occurs, we run the two tests (Standard Test, Customer Specific Test) on the online data to calculate the online risk probability $R_{online}$.

Our proposed algorithm *RISK* shows the steps involved in calculating $R_{online}$. The parameters for the algorithm are standard rules (SR), customer specific rules (CSR), Feature Scores (FS), and a batch of online transactions (T). Details of the rules, rule mappings, risk calculation steps, and flowcharts are described in detail in our previous work [3]. Each and every transaction is passed through the *StandardTest* function, which returns the violated standard rules (if any), which is then passed to the *CustomerSpecificTest* function to find out the valid causes for the breaking standard rules. There is a mapping between causes and the standard rules. Also different causes carry different weights based on the following criteria: 1) the mapping in between causes and standard rules, 2) the mapping between standard rules and features, and 3) the ranking of associated features. We use the term impact coefficient and weight interchangeably. To calculate risk probability from online data we use the following formula:

$$R_{Online} = [\ 1 - \frac{\Sigma\ \text{Impact Coefficient}\ (X)}{\Sigma\ \text{Impact Coefficient}\ (Y)}\ ] \times 100$$

Here, X is the set of valid causes for breaking the standard rules, and Y is the set of relevant causes (valid or invalid) for breaking the standard rules. Some of the use cases of the formula are as follows: 1) there is no valid cause for breaking a standard rule given X is equal to null and the $R_{Online}$ becomes maximum, 2) there are *n* related causes and all causes are valid given both X and Y are equal and the $R_{Online}$ becomes zero, and 3) there are *m* valid causes among *n* related causes given $R_{Online}$ is a value in between maximum and minimum. Thus, our proposed *RISK* algorithm returns the online risk probability $R_{online}$ for a transaction.

---

RISK (SR, CSR, FS, T):

1. for each online transaction **t** of **T**
2.     VioletedRules ← StandardTest(t)
3.     if count of ViolatedRules is greater than 0
4.         $R_{online}$ ← CustomerSpecificTest (ViolatedRules)
5.     else
6.         $R_{online}$ ← 0
7. return $R_{online}$

---

Finally, the risk probability from both *online* and *offline* data are combined using a weighted method to see whether the account is going to default in the near future.

## IV. DATA

In this work, we have used the "Taiwan" dataset [15] of Taiwan's credit card clients' default cases which has 23 features and 30,000 instances, out of which 6,626 (22.1%) are default cases. The same dataset has also been used in other research work [1][2]. Some of the features of this dataset are credit limit, gender, marital status, last 6 months bills, last 6 months payments, and last 6 months re-payment status. Records are labeled as either 0 (non-default) or 1 (default). Fig. 1 shows a snapshot of 5 random records from the dataset.

As indicated earlier, the *Heuristic Approach* processes two different datasets related to credit card transactions: the *offline* data and the *online* data. However, one of the issues with research in this area is that both offline *and* online transactional data are not publicly available. Specifically, there are some public datasets that contain customer summarized profile information and credit information, but not individual credit transactions. (i.e., no single publicly available dataset that contains both for the same set of customers). In order to tackle this issue, and provide a relevant data source for future work in this area (something that we will make publicly available after publication), we will decompose the Taiwan dataset into both *offline* and *online* datasets as shown with the examples in Table 1 and Table 2.

| ID | LIMIT_BAL | SEX | EDUCATION | MARRIAGE | AGE | PAY_0 | PAY_2 | PAY_3 | PAY_4 | ... | BILL_AMT4 | BILL_AMT5 | BILL_AMT6 | PAY_AMT1 | PAY_AMT2 | PAY_AMT3 | PAY_AMT4 | PAY_AMT5 | PAY_AMT6 | default payment next month |
|---|---|---|---|---|---|---|---|---|---|---|---|---|---|---|---|---|---|---|---|---|
| 18427 | 150000 | 2 | 2 | 1 | 24 | -1 | -1 | -1 | -1 | ... | 1596 | 405 | 0 | 902 | 177 | 1596 | 405 | 0 | 0 | 0 |
| 4615 | 180000 | 1 | 3 | 1 | 27 | 0 | 0 | 0 | 0 | ... | 5891 | 753 | 21753 | 2000 | 1700 | 300 | 0 | 21000 | 1000 | 1 |
| 16020 | 360000 | 2 | 2 | 2 | 31 | -2 | -2 | -2 | -2 | ... | -7 | -7 | 3500 | 2668 | 3000 | 7 | 0 | 3507 | 2500 | 0 |
| 4283 | 390000 | 1 | 1 | 1 | 35 | 0 | 0 | 0 | 0 | ... | 49414 | 51380 | 50661 | 20000 | 5006 | 5006 | 5008 | 3004 | 3006 | 0 |
| 16973 | 450000 | 1 | 1 | 1 | 67 | -2 | -2 | -2 | -2 | ... | 0 | 0 | 0 | 0 | 0 | 0 | 0 | 0 | 0 | 1 |

Fig. 1. Taiwan dataset

TABLE 1. OFFLINE DATASET CREATED FROM TAIWAN DATASET

| account | balance_limit | sex | education | marriage | age | total_bill | total_payment | repayment | default |
|---|---|---|---|---|---|---|---|---|---|
| 4663 | 50000 | 2 | 3 | 2 | 23 | 28718 | 1028 | 0 | 0 |
| 13181 | 100000 | 2 | 3 | 2 | 49 | 17211 | 2000 | 0 | 0 |
| 21600 | 50000 | 2 | 2 | 2 | 22 | 28739 | 800 | 0 | 0 |
| 1589 | 450000 | 2 | 2 | 2 | 36 | 201 | 3 | -1 | 0 |
| 28731 | 70000 | 2 | 3 | 1 | 39 | 133413 | 4859 | 2 | 0 |

TABLE 2. ONLINE DATASET CREATED FROM TAIWAN DATASET

| tid | account | amount | date | type |
|---|---|---|---|---|
| 53665 | 23665 | 660 | 2015-05-29 | pay |
| 9328 | 9328 | 46963 | 2015-05-14 | exp |
| 37597 | 7597 | 3000 | 2015-05-29 | pay |
| 9495 | 9495 | 75007 | 2015-05-14 | exp |
| 34113 | 4113 | 5216 | 2015-05-29 | pay |

Initially, from each record (customer) in the Taiwan dataset, we created 5 *online* transactions of type "pay" (payment) from PAY_AMT1 to PAY_AMT5 and 5 *online* transactions of type "exp" (expenditure) from BILL_AMT1 to BILL_AMT5. Since BILL_AMT is the sum of all individual bills or transactions, we divided this BILL_AMT into individual transactions by following the data distribution of a real credit card transactions dataset. As shown in Table 3, BILL_AMT1 is the total bill and PAY_AMT1 is the payment amount for the month of September 2005, BILL_AMT2 is the total bill and PAY_AMT2 is the payment amount for the month of August 2005, and so on, up to the oldest month, which in this case is April 2005 (BILL_AMT6 and PAY_AMT6). So, initially PAY_AMT6 and BILL_AMT6 go into the *total_payment* and *total_bill* for the *offline* data (Table 1). At the end of the month, the *total_payment* and *total_bill* is updated with that month's total bill (BILL_AMT) and total payments (PAY_AMT).

TABLE 3. MONTH VS FEATURE MAPPING IN TAIWAN DATASET

| Month | Feature ( BILL AMOUNT ) | Feature ( PAYMENT AMOUNT ) |
|---|---|---|
| April | BILL_AMT6 | PAY_AMT6 |
| May | BILL_AMT5 | PAY_AMT5 |
| June | BILL_AMT4 | PAY_AMT4 |
| July | BILL_AMT3 | PAY_AMT3 |
| August | BILL_AMT2 | PAY_AMT2 |
| September | BILL_AMT1 | PAY_AMT1 |

As stated previously, BILL_AMT is the summarized information of an entire month's transaction. We then break down this BILL_AMT into the individual transactions by following the real credit card transaction data distribution of the "Spain" dataset [16] used in the work [18]. We then scaled those datasets up/down as needed to convert them into the same currency scale using the formula below:

$$V2 = \frac{(Max2 - Min2) \times (V1 - Min1)}{(Max1 - Min1)} + Min2$$

where, V2 = converted value, Max2 = the ceiling of the new range, Min2 = the floor of the new range, Max1 = the ceiling of the current range, Min1 = the floor of the current range, V1 = the value needs to be converted.

To ensure that a corresponding transaction distribution can be followed in the "Spain" dataset for a BILL_AMT in the "Taiwan" dataset, we used *equal frequency binning* to determine the ranges under which a monthly bill amount (BILL_AMT) must fall into. *Equal frequency binning* uses an inverse cumulative distribution function (ICDF) to calculate the upper and lower ranges. As a result, we came up with on average 359,583 online transactions per month for the same 30,000 accounts or records in the original dataset.

It should also be noted that another significant result of this work is the creation of a dataset for other researchers. As mentioned earlier, public access to credit card summary data and credit card transactional data for the same set of customers is rare. While it was necessary to create this dataset for our specific research purposes, we realize the benefit of making this dataset public to the general research community.

V. EXPERIMENT

For our experiments, we will use the Python *scikit-learn* library. The following sections describe the experimental setup for each of the two approaches that we discussed earlier.

*A. Machine Learning Approach*

We will run different machine learning algorithms on the "*Taiwan*" dataset. The purpose of this test is to evaluate an improved approach in terms of the following performance evaluation metrics: accuracy, recall, F-score, and precision. We chose these metrics for two reasons: 1) these are the metrics frequently used in related research, and 2) to compare the results with previous research using this Taiwan dataset.

Some of the algorithms we tried include K-nearest Neighbor, Random Forest, Naïve Bayes, Gradient Boosting, Extremely Random Trees (Extra Trees), etc. We also used the k-fold (k =10) cross-validation technique for the testing/training set split and to calculate performance metrics. Default parameters for all algorithms (in scikit-learn) were used unless otherwise mentioned.

*B. Heuristic Approach*

This approach originated from our previous preliminary work using a synthetic dataset [3]. However, in this work, we will validate our approach by using the publicly available "Taiwan" dataset, and dividing the dataset into *offline* and *online* datasets. Beside solving the limitations (e.g., lack of validating the proposed model using a known and real dataset), we also found a better base algorithm (Extremely Random Trees) than before that will contribute to the calculation of the offline risk probability $R_{offline}$ in our *Heuristic Approach*. We briefly reiterate the two tests as discussed in detail in [3]:

*1) Standard Test:* The purpose of this test is to identify transactions that deviate from the normal behavior and pass them to the next test named *Customer Specific Test.* Here the normal behavior refers to the common set of standard rules that every good transaction bound to follow. Some of the standard rules that we applied to the "Taiwan" dataset include whether the minimum due was paid, whether the paid amount was less than the bill amount, whether the monthly total bill was greater less than balance limit etc.

*2) Customer Specific Test:* This test is more customer-centric rather than the standard rules of *Standard Test* that are applicable to every account in the same way. It takes customer specific measures like foreign national, job change, address change, promotion, salary increase, etc. into consideration. The purpose of this test is to recognize possible causes for which a transaction is unable to satisfy a standard rule in the *Standard Test*. In the experiment with the *"Taiwan"* dataset, this test was not completely in effect due to the lack of necessary information that can be extracted from the dataset.

As a consequence of the above tests, an online risk probability $R_{online}$ is returned from the *RISK* algorithm explained earlier in section III. Details of this procedure are described in our previous work [3]. The total risk probability for a transaction comes from both *online* and *offline* data. So, the equation of total risk probability is as follows:

$$R_{Total} = R_{Online} + R_{Offline} \quad (1)$$

Here,

$R_{Total}$ = Overall risk probability from both *online* and *offline* data.

$R_{Online}$ = Risk probability from *online* data

$R_{Offline}$ = Risk probability from *offline* data

Initially, we get the risk probability from *offline* data ($R_{Offline}$) for corresponding accounts from the value of the risk probability distribution value of the classification results on *offline* data. So, for the first transaction of the account, the $R_{Offline}$ = *probability of it being defaulted* comes from the probability distribution of the classification outcome. Then, for the subsequent transaction N, the $R_{offline}$ is the value of total risk probability of from the previous transaction.

$$R_{Offline} \text{ of transaction } N = R_{Total} \text{ of transaction } N-1$$

Thus, $R_{offline}$ is updated in two situations: a) At the end of a transaction that has a positive $R_{online}$; and b) At the end of the month to synchronize with possible profile changes (i.e., credit limit increase). That is the reason why we created 5 batches from 5 months of data and ran them chronologically to accommodate the profile change at the end of each month, which also leads to better results (Table 4 and Fig. 4).

Furthermore, the risk probability from the online data and offline data may carry different weights. For example, giving half of the weight (i.e., 50%) to offline data and remaining half of the weight (i.e., 50%) to online data might provide better mining results for a particular company or dataset. On the other hand, for another company or dataset, a different combination of offline vs online risk probability weights might be better. So, the modified version of (1) for a total risk probability calculation is:

$$R_{Total} = \lambda R_{Online} + (1-\lambda) R_{Offline} \quad (2)$$

where λ is the risk factor.

For our experiments, we have found that between 45% and 50% for the *online* data weight, with the remaining % for the *offline* data weight, provides the best results. In other words, if λ = .45 or .5 then 1- λ = .55 or .5 accordingly. As a result, we have used λ = .5 for our experiments.

## VI. RESULTS

Running the different algorithms on the "Taiwan" dataset we discover that the *Extremely Random Trees* outperforms all the standard machine learning algorithms and notable previous works [1][2] on this dataset in terms of Accuracy, Precision, Recall and F-score. Detail scores are shown in Fig. 2. The performance gain is mainly due to the fact that the tree-based approach works very well for problems where the number of features is moderate, data is properly labeled, and there are few missing values. To the best of our knowledge, the Extremely Random Trees algorithm has not been used on this dataset before.

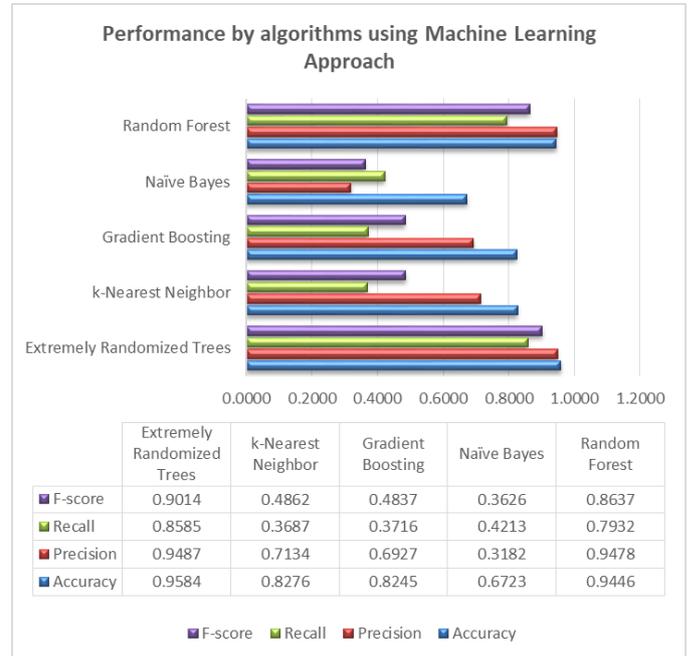

|  | Extremely Randomized Trees | k-Nearest Neighbor | Gradient Boosting | Naïve Bayes | Random Forest |
|---|---|---|---|---|---|
| F-score | 0.9014 | 0.4862 | 0.4837 | 0.3626 | 0.8637 |
| Recall | 0.8585 | 0.3687 | 0.3716 | 0.4213 | 0.7932 |
| Precision | 0.9487 | 0.7134 | 0.6927 | 0.3182 | 0.9478 |
| Accuracy | 0.9584 | 0.8276 | 0.8245 | 0.6723 | 0.9446 |

Fig. 2 Performance by algorithms using Machine Learning Approach

Fig. 3 shows the comparison of performance using the *Machine Learning Approach* (i.e., applying only the best classifier, Extremely Random Trees on the dataset without any other test like *Standard Test* or *Customer Specific Test*), the *Heuristic Approach*, and the State-of-the-art. So far, we have seen a maximum accuracy of 84% (82% on training data) and maximum recall of 65.54% among all previous research work on this "Taiwan" dataset, while the *Heuristic Approach* has an accuracy of 93.14% and the *Machine Learning Approach* has an accuracy of 95.84%. We also realize a better recall percentage.

In fraud or risk, detection recall is very important because we don't want to miss fraud or risks. However, maximizing recall introduces an increase of False Positives, which is expected in risk analytics.

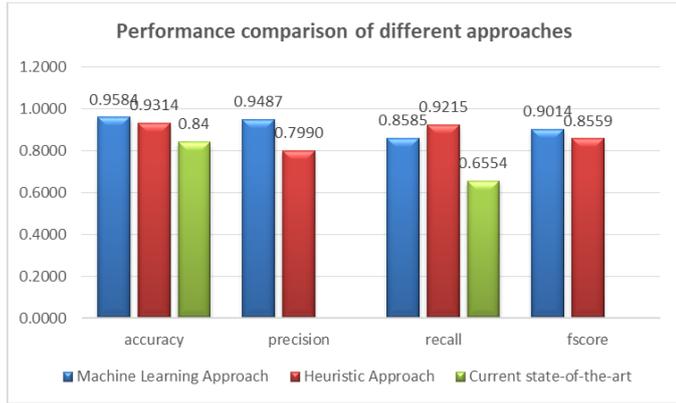

Fig. 3 Performance comparison of different approaches

Recall that we divided the dataset into *offline* and *online* datasets, as mentioned in the Data section (Section IV), consisting of bill and payment data for 6 months. Data from the first month was included in the summarized fields (i.e., total_bill and total_payment), and for the remaining 5 months, we made 5 batches of *offline* data and 5 batches of *online* data. We then ran *offline* batch 1 and *online* batch 1 serially, followed by *offline* batch 2 and *online* batch 2, and so on, up to batch 5, which leads to comparing the results from the newly created offline (Table 1) and online (Table 2) dataset from the "Taiwan" dataset with the results of the *Machine Learning Approach*, as shown in Fig 3.

TABLE 4. BATCH WISE PERFORMANCE METRICS

| Batch | Accuracy | Precision | Recall | F-score | Computation Time (offline) | Computation Time (online) |
|---|---|---|---|---|---|---|
| 1 | 0.94 | 0.92 | **0.82** | 0.87 | 12.23 | 152.54 |
| 2 | 0.94 | 0.88 | 0.86 | 0.87 | 9.14 | 109.35 |
| 3 | 0.94 | 0.84 | 0.89 | 0.87 | 13.04 | 86.90 |
| 4 | 0.94 | 0.82 | 0.91 | 0.86 | 11.59 | 89.74 |
| 5 | 0.93 | 0.80 | **0.92** | 0.86 | 10.21 | 133.58 |

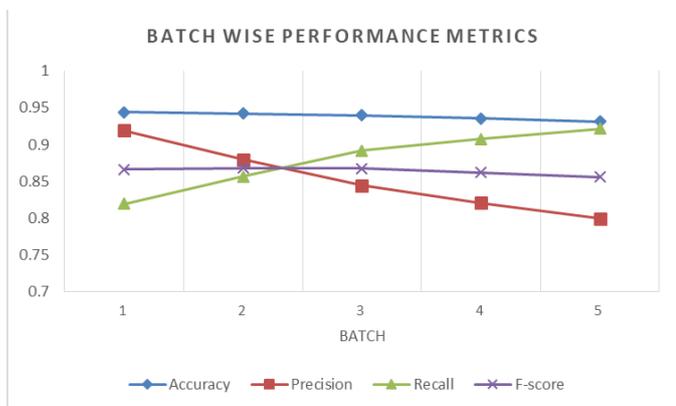

Fig. 4. Batch wise performance metrics

From Table 4 and Fig. 4 we can see that recall increases as the number of batches increase. This implies that the percentage of targeted (i.e., defaulting) accounts increases (linearly) with the number of batches (i.e., as more information about the customer is known).

The computation time for both the *Machine Learning Approach* and calculating $R_{offline}$ using *Extremely Random Trees* for 30,000 accounts was on average 11.24 seconds using a commodity laptop with an Intel core i7 processor and 12 GB RAM. Though *Naïve Bayes* is a bit faster than *Extremely Random Trees*, its performance in terms of accuracy, precision, recall, and F-score are not. For the online data computation, it took on average of 114.42 seconds for a batch size of on average of 359,583 transactions. For our interpretation of results, we created only one batch per month. However, there is nothing in our proposed approach that requires batches of this size, and any number of transactions per month for online data could be used, which could lead to batches with a much smaller number of transactions with less computation time. To verify this, we tried with batches of different sizes (reducing the batch size by half each time) and we found that the computation time for the online data reduces almost linearly with the reduction of the number of transactions per batch. From Fig. 5, we can see that the trendline (dotted line) is almost in line with the actual line. This demonstrates how fast this approach can process the *online* transaction and give a decision in near real-time.

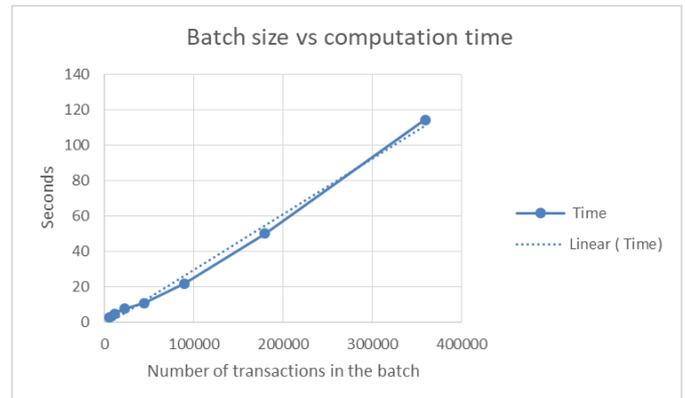

Fig. 5 Batch size vs computation time

Another mentionable contribution of this approach is the early detection. While there is ~10% improvement in recall from the first month (batch 1) to the fifth month (batch 5) – from a recall of 81.96% to 92.15% - it is clear that we can achieve a good recall very early in the process, enabling a real-time system to detect potential credit card default.

## VII. CONCLUSION

In this research, we have used two approaches, *Machine Learning* and *Heuristic*, for mining default accounts from a well-known dataset. The *Heuristic Approach* came from our previous work [3], that we validated with actual data in this work. The main idea of the *Heuristic Approach* is to calculate the risk factor from the recent transactional data (*online*) and combine the results with pre-computed risk factors from historical (*offline*) data in an efficient way. To make the process efficient, we only have to process a transaction when it initially occurs, and then the combined risk factor is carried forward for future transactions. We showed this approach can predict a default account significantly in advance, which is very cost efficient for the funding organization. In addition, we demonstrated that the performance of both approaches outperforms reported approaches using the same data set [1][2]. Our future plan is to improve the *Heuristic Approach* so that it outperforms the *Machine Learning Approach* in terms of all performance metrics, and validate that with multiple datasets. Other plans include: testing and validating the model with multiple real datasets, standardizing the online vs offline risk weight ratio (the value of λ) with multiple datasets of credit defaults, as well as handling concept drift to deal with change in the distribution of the online data over time which may affect the effectiveness of the approach.